\documentclass[letterpaper, 10 pt, conference]{ieeeconf}  % Comment this line out if you need a4paper

\usepackage{graphicx} % for pdf, bitmapped graphics files
\usepackage{amsmath} % assumes amsmath package installed
\usepackage{amssymb}  % assumes amsmath package installed
\usepackage{hyperref}
\usepackage{wrapfig}

\IEEEoverridecommandlockouts                              % This command is only needed if 
\title{\LARGE \bf Action sequencing using visual permutations}

\author{Michael~Burke$^{\dagger}$, Kartic~Subr$^{\ddagger}$ and Subramanian~Ramamoorthy$^{\ddagger}$% <-this % stops a space
\thanks{*This research was supported by the Alan Turing Institute, as part of the Safe AI for surgical assistance project, while M. Burke was at the University of Edinburgh.}% <-this % stops a space
\thanks{$^{\dagger}$Electrical and Computer Systems Engineering, Monash University, Australia. 
        {\tt\small michael.burke1@monash.edu}}
\thanks{$^{\ddagger}$School of Informatics, University of Edinburgh, United Kingdom.
        {\tt\small \{k.subr,s.ramamoorthy\}@ed.ac.uk}
        }}

\begin{document}

\maketitle
\thispagestyle{empty}
\pagestyle{empty}

\begin{abstract}
   Humans can easily reason about the sequence of high level actions needed to complete tasks, but it is particularly difficult to instill this ability in robots trained from relatively few examples. This work considers the task of neural action sequencing conditioned on a single reference visual state. This task is extremely challenging as it is not only subject to the significant combinatorial complexity that arises from large action sets, but also requires a model that can perform some form of symbol grounding, mapping high dimensional input data to actions, while reasoning about action relationships. This paper takes a permutation perspective and argues that action sequencing benefits from the ability to reason about both permutations and ordering concepts. Empirical analysis shows that neural models trained with latent permutations outperform standard neural architectures in constrained action sequencing tasks. Results also show that action sequencing using visual permutations is an effective mechanism to initialise and speed up traditional planning techniques and successfully scales to far greater action set sizes than models considered previously.
\end{abstract}

% \begin{IEEEkeywords}
% IEEE, IEEEtran, journal, \LaTeX, paper, template.
% \end{IEEEkeywords}

% \fi
% \IEEEpeerreviewmaketitle

\section{INTRODUCTION}

%===============================================================================
Humans possess a remarkable ability to plan and select actions to rearrange scenes and form infinitely many newly imagined constructs, relying on visual information to guide the process. Taking inspiration from this ability, this paper considers the challenge of learning to sequence a set of high level actions to solve a task depicted in a \textit{single} reference input image, using as few demonstrations as possible.

Specifically, we are interested in constrained action sequencing tasks where multiple actions can be selected to solve a task, but each action can only be chosen a finite number of times. Constraints like these are particularly common in robotics, and typically present in all assembly tasks. For example, assembling a stool may require that a seat is attached to four identical legs, using 4 identical screws. The symmetry present in this task means that we have multiple ways of performing this assembly, resulting in ambiguity in action sequencing. This work shows that existing neural action sequencing approaches fail in this setting, and introduces a model that, by construction, copes with this ambiguity.

Traditionally, action sequencing of this form has been the domain of planning and reasoning \cite{fox2003pddl2}, relying on pre-trained perception modules and known transition dynamics, with clearly defined symbolic rules and constraints. However, more recently, our community has focused on data driven methods relying on neural network universal function approximators to build policies trained using a range of mechanisms, from behaviour cloning \cite{pomerleau1989alvinn,torabi2018behavioral} to deep reinforcement learning \cite{mnih2015human,lillicrap2015continuous}. Much of our focus here has been on training or learning mechanisms, and there has been arguably less emphasis in robotics on the effect of the architectures we use and the inductive biases therein.

\begin{figure}
    \centering
    \includegraphics[width=0.45\textwidth]{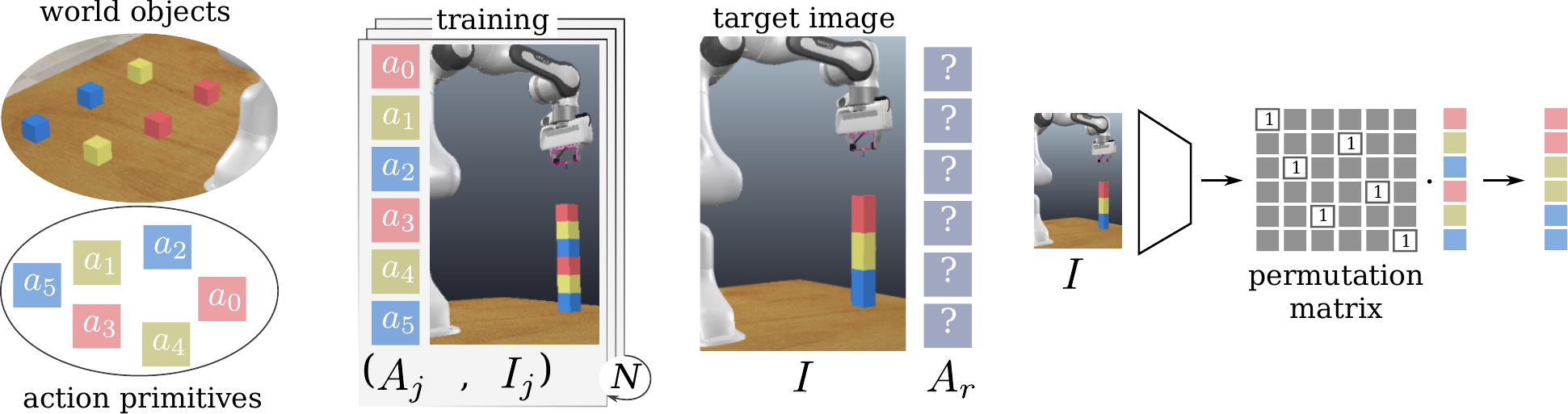}
    \caption{This paper considers the task of learning to predict a target sequence of actions $A_r$, given a \textit{single} reference input image $I$, from as few demonstrations as possible $\{A_j, I_j\}$. Looking at the towers above, it is easy to see that by rearranging the blocks on the left we can build the tower on the right. We apply this \textit{permutation} perspective to the problem of neural action sequencing.}
    \label{fig:concept}
\end{figure}

However, the choice of neural architecture strongly dictates the solutions we may find. For example, consider the tower building task in Figure \ref{fig:concept}, where our robot is required to select and place blocks to build the tower depicted in an input image. One approach to solving this task may be to consider tower building as a process of arbitrarily selecting blocks from an existing set (a fully connected neural classification model). Alternatively, we could frame tower building as a sequential process, where blocks selected are conditioned on previous block selections (a neural classification model with a temporal output layer). Unfortunately, neither of these approaches enforce a particularly important constraint that is common to classical planning systems \cite{Dantam-RSS-16}, but absent from modern neural models -- once a block has been placed, it can no longer be used in future. 

This common-sense constraint is obvious to humans, but it is unclear how to formulate a training objective that accounts for this, particularly if there is ambiguity arising when multiple actions can be associated with similar visual inputs. This paper explores a \textit{permutation} perspective of action sequencing, introducing a neural architecture with latent permutations that allows for constrained, variable-length action sequencing from high-dimensional pixel inputs. 

We investigate this model using a series of experiments conducted in a behaviour cloning setting, and show that while augmenting existing neural classification models with post-hoc symbolic constraints is reasonably effective at dealing with action re-use constraints in small scale settings, we gain significant improvements by directly embedding these into neural models using latent permutations. A particularly important finding of this paper is that action sequencing using latent permutations scales to significantly larger action set sizes than standard neural models, and copes well with combinatorially complex settings. In summary, the contributions of this paper are:
\begin{enumerate}
\item the formulation of constrained action sequencing as a problem of learning to permute a discrete set (Sec. \ref{sec:prob_form});
\item a latent permutation modelling approach that outperforms
standard neural models on vision-based action sequencing tasks (Sec. \ref{sec:model}, \ref{sec:baseline} and \ref{sec:scale});
\item the application of the latent permutation model to generalise to new
concepts using previously encountered action subsets (Sec. \ref{sec:unique} and \ref{sec:variable_length});
\item a demonstration of the potential performance gains that can be
obtained by using vision-based action sequencing to initialise
optimisation-based planning algorithms (Sec. \ref{sec:soma});

\end{enumerate}
 
\section{RELATED WORK} 

Robotics has traditionally made a distinction between higher-level symbolic or task-level planning, and control at a behavioural level \cite{konidaris2018skills}. The former typically assumes that a domain specific language (DSL) defining objects, predicates, actions or operations and goals is available. For example, in robotic assembly \cite{5390994,8629046,6630673}, the domain in which our work has most relevance, DSLs may include placement actions defined in a given object frame, with contact constraints and other pre and post conditions. Similarly, in carpentry planning this may include materials and parts, along with associated tools and operations that can be applied to these \cite{wu_siga19}. Here, planning is typically formulated as a constrained search or optimisation problem, which can quickly become computationally expensive in more complex task settings. Early approaches dealing with this search relied on linear programming-like possibility trees or knowledge graphs \cite{popplestone1980interpreter,Popplestone_Liu_Weiss_1990}, which try to prune the search space over actions by taking constraints into account. More recently, L{\'a}zaro-Gredilla et al. \cite{lazaro2019beyond} learn concepts as cognitive programs by searching for algorithms that could generate a demonstrated scene, but this approach is limited to very simple scenes due to the need for dedicated scene parsers. In more complex settings it can be particularly challenging and time-consuming to develop a DSL, and inferring states from partially-observable uncertain environments is non-trivial.

In contrast, data-driven approaches like behaviour cloning \cite{pomerleau1989alvinn} or deep reinforcement learning \cite{mnih2015human} try to avoid the need to specify a DSL or carefully program a robot, generally relying on universal neural network function approximators and substantial amounts of training data \cite{lake2017building} to produce suitable policies that act directly on high dimensional observations. These connectionist models fail to incorporate many of the symbolic or logical constraints that are typically present in robot task planning settings. Existing attempts to extend neural models to handle these types of constraints are often made in a post-hoc fashion (eg. action clipping \cite{DBLP:conf/icml/FujitaM18}, elaboration using auxiliary losses \cite{Innes-RSS-20}). In this paper, we incorporate symbolic planning ideas around symmetries and permutations, which are often exploited in constraint programming to speed up search \cite{fox2002extending,GENT2006329,pochter2011exploiting}, into neural models through the use of latent permutations. In so doing, we gain generalisability and an improved ability to handle ambiguities in action selection at scale. 

Connectionist policies and model-based symbolic planning systems are by no means incompatible.
As a practical middleground, there has been increasing interest in pruning the search space to speed up planning by using neural networks as universal function approximators. For example, Dreamcoder \cite{Ellis2020DreamCoderGG} relies on a neural model to propose suitable program structures to speed up search in a program induction setting. A similar technique has been used to interpret transition system dynamics, iteratively refining a priority queue of candidate solutions \cite{penkov2017using}. Neural surrogate modelling has also been broadly applied to warm start general purpose optimisation procedures \cite{banerjeelearning,velazco2002neural}. Along these lines, and closest to this work, Driess et al. \cite{Driess-RSS-20} propose an image conditioned recurrent model to predict sequences of up to 6 actions, which are then used to speed up symbolic robot planning. Our approach is similar, but, as shown in this work, action sequencing with latent permutations explicitly allows for learning in the presence of action constraints, and significantly outperforms temporal models in settings where action ambiguity may exist and action set sizes are scaled.

Deep learning using latent permutations is a recent approach that has proven useful in differentiable sorting and ranking applications \cite{cuturi2019differentiable,blondel2020fast,mena2018learning, paulus2020gradient}, and in computer vision for a range of applications including semi-supervised learning \cite{mena2018learning,santa2017deeppermnet}, captioning \cite{Cornia_2019_CVPR} and point cloud segmentation \cite{Rosu-RSS-20}. However, to the best of our knowledge, this work is the first to explore their use for sequence modelling.

\section{PRELIMINARIES}

\subsection{Problem formulation}
\label{sec:prob_form}
This work considers a behaviour cloning setting, where we are required to learn an open-loop, image-conditioned action sequencing policy from demonstrations. More formally, assume that a robot is required to correctly order N action primitives, $a_i \in \mathbf{A}=\{a_1 \hdots a_N\}$, to accomplish some task described by an image $I$, depicting a reference state associated with the task. Our goal is to use behaviour cloning to learn to predict an action sequence $\mathbf{A}_r$  that will reproduce (or deconstruct) the scene depicted in a query image $I$ using prior training examples comprising $M$ action sequences and reference images $(\mathbf{A}_j,I_j)$, $j~\in~\{1, \ldots, M\}$.

\subsection{Baselines: Action sequencing using behaviour cloning and Hungarian assignment}
A naive approach to addressing the problem above would be to train a multi-class, multi-label feed-forward convolutional neural network $\mathbf{X} = g_\theta(I)$ with parameters $\theta$, to predict action sequences directly, using a cross-entropy classification loss,
\begin{equation}
\mathcal{L} = - \frac{1}{N}\sum_{i=1}^N\sum_{j=1}^N y_{i,j} \log x_{i,j}.\label{class_loss}
\end{equation}

Here, $y_{i,j}$ is a binary label indicating the use of action $a_j$ in the $i$-th step of the demonstrated action sequence, while $x_{i,j} \in \mathbf{X}$ is a logit predicted by the neural network.

Since action sequencing is crucial to obtain a desired goal state, failure to predict even a single action correctly will result in failure to complete the task. This makes action sequencing using the behaviour cloning approach described above particularly challenging. Moreover, this naive approach to behaviour cloning is limited as there are no constraints on the model preventing action re-use. This poses problems if action ambiguity is present. For example, in Fig. \ref{fig:concept} there are two blocks of each colour, so it is possible that a model lacking the ability to reason about objects or actions already performed would attempt to call actions to pick and place the same object twice when attempting assembly. 

\begin{figure}
    \centering
    \includegraphics[width=0.35\textwidth]{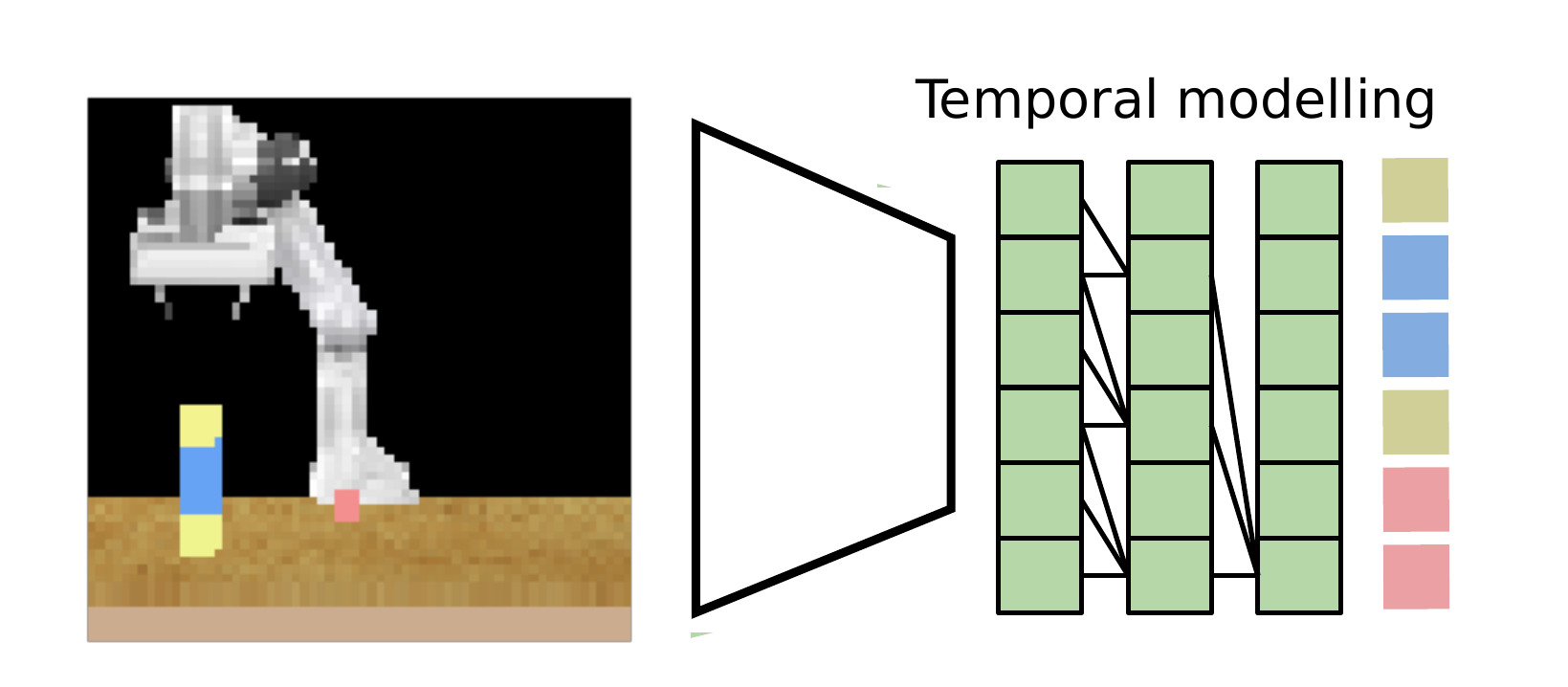}
    \caption{Sequence modelling can capture temporal ordering information in action sequences, but does not explicitly prevent action re-use.}
    \label{fig:TCN}
\end{figure}
A standard approach to incorporating temporal information like this is to rely on sequence modelling, using recurrency \cite{hochreiter1997long,graves2013generating} or  temporal convolutions \cite{bai2018empirical,lea2017temporal} in the output sequence prediction, as illustrated in Fig. \ref{fig:TCN}. Models like these have recently been proposed for vision-based action sequencing \cite{Driess-RSS-20}. However, these models do not explicitly incorporate constraints on action re-use, which are common in robotics. The ability to reason about permutations is valuable across a wide range of tasks in robotics, and often studied as a balanced linear assignment problem, with the goal of identifying a permutation or assignment matrix $\mathbf{P}$ that remaps some standard ordering $\mathbf{A}_o$, 
\begin{equation}
\mathbf{A}_r = \mathbf{P} \mathbf{A}_o
\end{equation}
so as to minimise some assignment cost $\sum_{a \in \mathbf{a}_o}  C_{a_o,a_r}$. The Hungarian algorithm \cite{kuhn1955hungarian,munkres1957algorithms} is a well known technique to solve problems of this form in polynomial time. By using the logits of the network $g_\theta(I)$ to produce an assignment cost, $C_{i,j} = 1 - x_{i,j}$, we can apply the Hungarian algorithm to order actions, and avoid issues around action re-use. However, if the classifier is overconfident in prediction (for example, predicts the same action twice with high probability), this assignment operation could introduce additional errors.

Instead of applying the Hungarian algorithm as a post-hoc assignment stage, it is natural to consider the possibility of learning with embedded inductive biases for action assignment. Recent approaches to differentiable sorting and ranking \cite{cuturi2019differentiable,blondel2020fast,mena2018learning, paulus2020gradient} provide a useful mechanism to learn about permutations in this manner.

\section{ACTION SEQUENCING USING SINKHORN NETWORKS}
\label{sec:model}
Differentiable sorting networks typically rely on the Sinkhorn operator $S(\mathbf{X})$, \cite{adams2011ranking,mena2018learning}, acting on a square matrix $\mathbf{X}$,  \begin{eqnarray}
S^0(\mathbf{X}) &=& \exp(\mathbf{X})\\
S^l(\mathbf{X}) &=& \mathcal{T}_{col}(\mathcal{T}_{row}(S^{l-1}(\mathbf{X})))\\
S(\mathbf{X}) &=& \lim_{l\to\infty}S^{l}(\mathbf{X})).
\end{eqnarray}
Here, $\mathcal{T}_{col}(\cdot)$ and $\mathcal{T}_{row}(\cdot)$ denote column and row normalisation operations respectively. Mena et al. \cite{mena2018learning} show that a differentiable approximation to the permutation $\mathbf{P}$ can be obtained using the Sinkhorn operator, 
\begin{equation}
\mathbf{P} = \lim_{\tau\to0^+} S(\mathbf{X}/\tau),
\end{equation}
with $\mathbf{X} = g_\theta(\cdot)$ a square matrix predicted using a suitable feed-forward neural network. Intuitively, this soft assignment operation can be thought of as the permutation analogue of a softmax operation.

The Sinkhorn operation \cite{adams2011ranking}, applied to a square matrix, repeatedly iterates between column and row-wise normalisation operators, so that the matrix resulting from this operation is a doubly stochastic (rows and columns sum to one) assignment matrix from the Birkhoff polytope. The extreme points of this polytope are permutation matrices. We can control how soft this assignment is by adding noise and adjusting a temperature parameter, $\tau$. This is important to allow the backpropagation of errors in assignment through this operation and down a neural network.

\subsection{Image conditioned action sequencing}

We make use of Sinkhorn networks to sequence robot actions by training a feedforward convolutional neural network to predict matrix $\mathbf{X}  = g_\theta(\mathbf{I})$, using the Sinkhorn operator (with Gumbel-Matching \cite{mena2018learning} to determine permutation $\mathbf{P}(g_\theta(\mathbf{I}))$. This network can be trained to minimise a mean squared error loss between sequenced actions,
\begin{equation}
\mathcal{L} = ||\mathbf{a}_j - \mathbf{P}(g_\theta(\mathbf{I}))\mathbf{a}_o||. \label{perm_loss}
\end{equation}
Here, $\mathbf{a}_o$ denotes a one-hot encoded base action sequence order, and $\mathbf{a}_j$ a one-hot encoded demonstrated action sequence sampled from the training set. At test time, when a soft approximation is no longer needed, action sequencing occurs by predicting a permutation matrix, and using the Hungarian algorithm for hard assignment.

By construction, a permutation is unable to re-use an action\footnote{Actions that are required more than once can be dealt with by adding additional instances of these to the action set.}, which forces the network to learn to deal with action ambiguities. Our hypothesis is that behaviour cloning models trained with explicit inductive biases towards permutations will be better suited to constrained action sequencing than feedforward and temporal convolutional neural networks.

\subsection{Coping with action subsets}
\label{sec:stopping}

None of the approaches described above are able to deal with restricted subsets of actions. For example, building a tower using 3 blocks does not require all actions be used, but the models above all assume that a fixed number of actions are required to complete tasks. 

\begin{figure}
    \centering
    \includegraphics[width=0.45\textwidth]{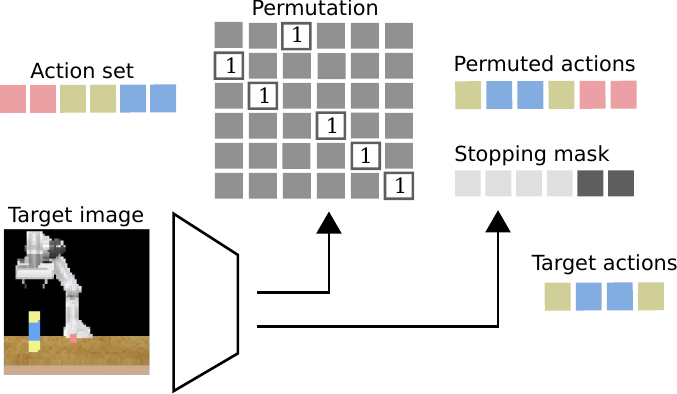}
    \caption{Framework for action sequencing using visual permutations.} 
    \label{fig:architecture}
\end{figure}
We extend the model above to handle action subsets using an auxiliary stopping network $f_\phi(\mathbf{I})$ parametrised by $\phi$, that predicts the number of actions required to complete a task for a given image. This network is trained using a standard cross-entropy classification loss.

The extension to subsets requires that we modify the loss in (\ref{perm_loss}) to allow for variable action sequence lengths. We accomplish this by masking the predicted and ground truth sequences in the respective loss functions. Fig. \ref{fig:architecture} provides an overview of the proposed action sequencing model. A Sinkhorn network is used to predict permutations over action sequences conditioned on a reference scene, and a masking network restricts the sequence of actions to only the subset required to complete the referenced task.

\section{EXPERIMENTAL RESULTS}
\label{sec:result}

We start by investigating the effects of including inductive biases for permutations in the neural network used for behaviour cloning for our running tower stacking example.

\subsection{Fixed length action sequencing}
\label{sec:baseline}
\begin{figure}
    \centering
    \includegraphics[width=0.4\textwidth]{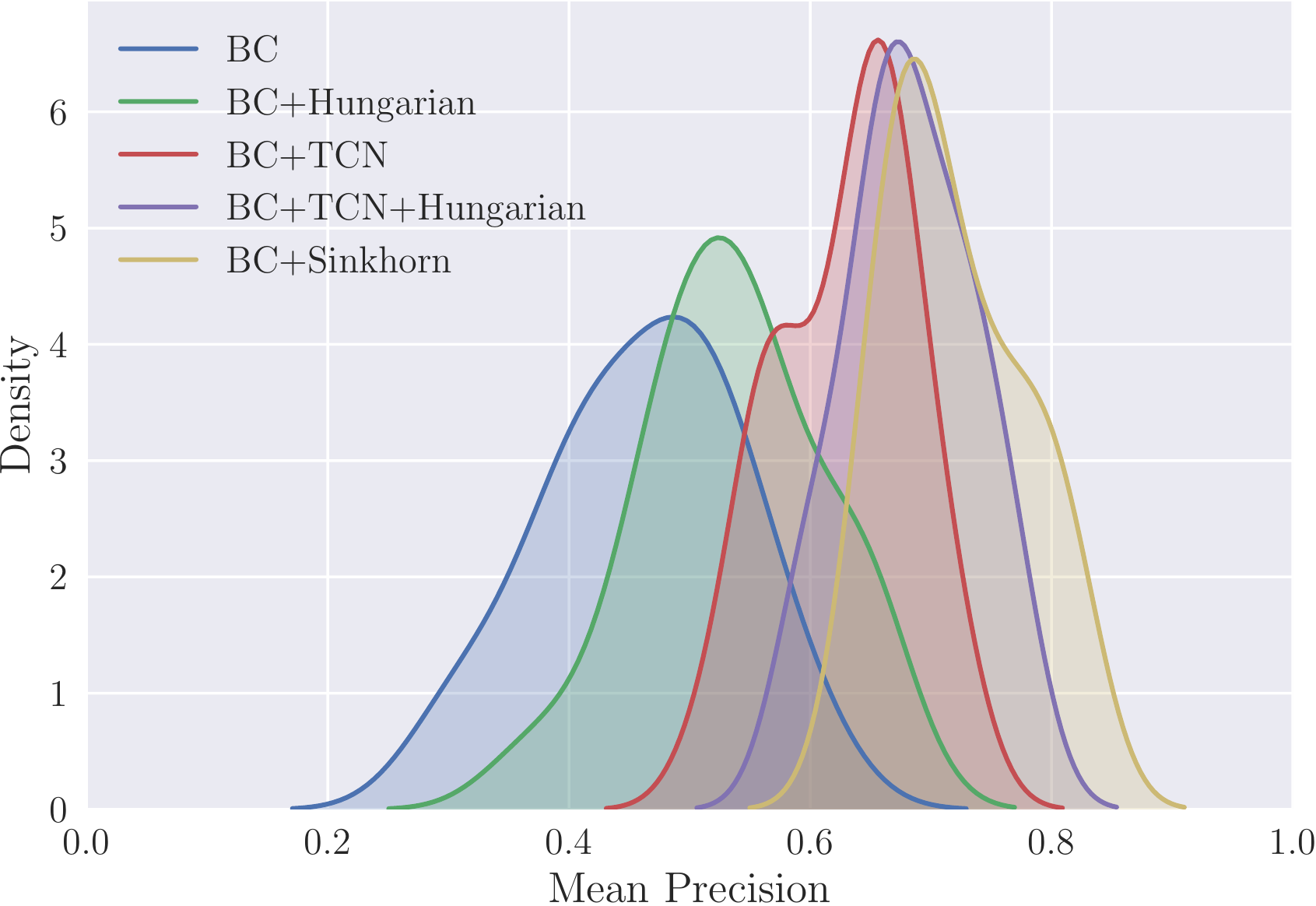}
    \caption{Mean average precision (correct block colour) distributions show that BC+Sinkhorn substantially outperforms behaviour cloning models that do not explicitly account for permutations.}
    \label{fig:BC_results}
\end{figure}
Here, robot actions consist of picking up a block from a known location and placing it on the previously placed block. Once a block has been placed, this action should no longer be called, as doing so would demolish the tower. We collect 300 tower stacking demonstrations with randomly ordered action sequences of length 6 using CoppeliaSim and PyRep \cite{james2019pyrep}, and save the corresponding image of the completed tower (see Fig. \ref{fig:concept}). 

We train on 200 demonstrations and evaluate on 100 held out demonstrations, using the mean average precision (number of times a block of the correct colour was correctly selected for placement) between predicted and ground truth action sequences. This experiment is repeated 20 times for models trained using different random seeds.

Models compared include direct behaviour cloning (BC) with a fully connected output layer, the same behaviour cloning network with post-hoc assignment using the Hungarian algorithm (BC+Hungarian), and action sequencing using Sinkhorn networks (BC+Sinkhorn). In addition, we also investigate behaviour cloning using a temporal convolutional neural network decoder (BC+TCN/ BC+TCN+Hungarian), a state-of-the art sequence modelling approach commonly used for action recognition \cite{lea2017temporal}, which enforces temporal structure in the predicted output sequence.

Figure \ref{fig:BC_results} shows kernel density estimates over the mean average precisions in block predictions over multiple seeds, while Table \ref{tab:BC_results} shows the average number of times a block selection action was re-used (The tower collapses upon action re-use). The inductive bias towards permutation prediction introduced using the Sinkhorn networks clearly improves action sequencing. 
\begin{table}
    \centering
    \caption{Block colour precision and tower building success}
    \begin{tabular}{|l|c|c|}
    \hline
         &  \bf{Colour Precision} & \bf{Action repetitions}\\
         & \bf{(Mean, Std. Dev.)} & \bf{(Tower Collapses)}\\
         \hline
         \hline
        \bf{BC} & $0.46 \pm 0.08$ & 46 \%\\
        \hline
        \bf{BC+Hungarian} & $0.53 \pm 0.07$ & 0 \% \\
        \hline
        \bf{BC+TCN} & $0.63 \pm 0.05$ & 53 \%\\
        \hline
        \bf{BC+TCN+Hungarian} & $ 0.68 \pm 0.05$ & 0 \%\\
        \hline
        \bf{BC+Sinkhorn} & $0.72 \pm 0.05$ & 0 \%\\
        \hline
    \end{tabular}
    \vspace{-5mm}
    \label{tab:BC_results}
\end{table}
While augmenting TCN and BC networks with post-hoc Hungarian algorithm assignment remedies problems where the same action is selected multiple times, and provides substantial improvement over direct behaviour cloning, both struggle with image conditioned tower building. Explicitly modelling temporal action sequence behaviours using TCNs improves performance, but is still outperformed by BC+Sinkhorn.

\subsection{Generalisation to unseen configurations}
\label{sec:unique}
\begin{figure}
    \centering
    \includegraphics[width=0.46\textwidth]{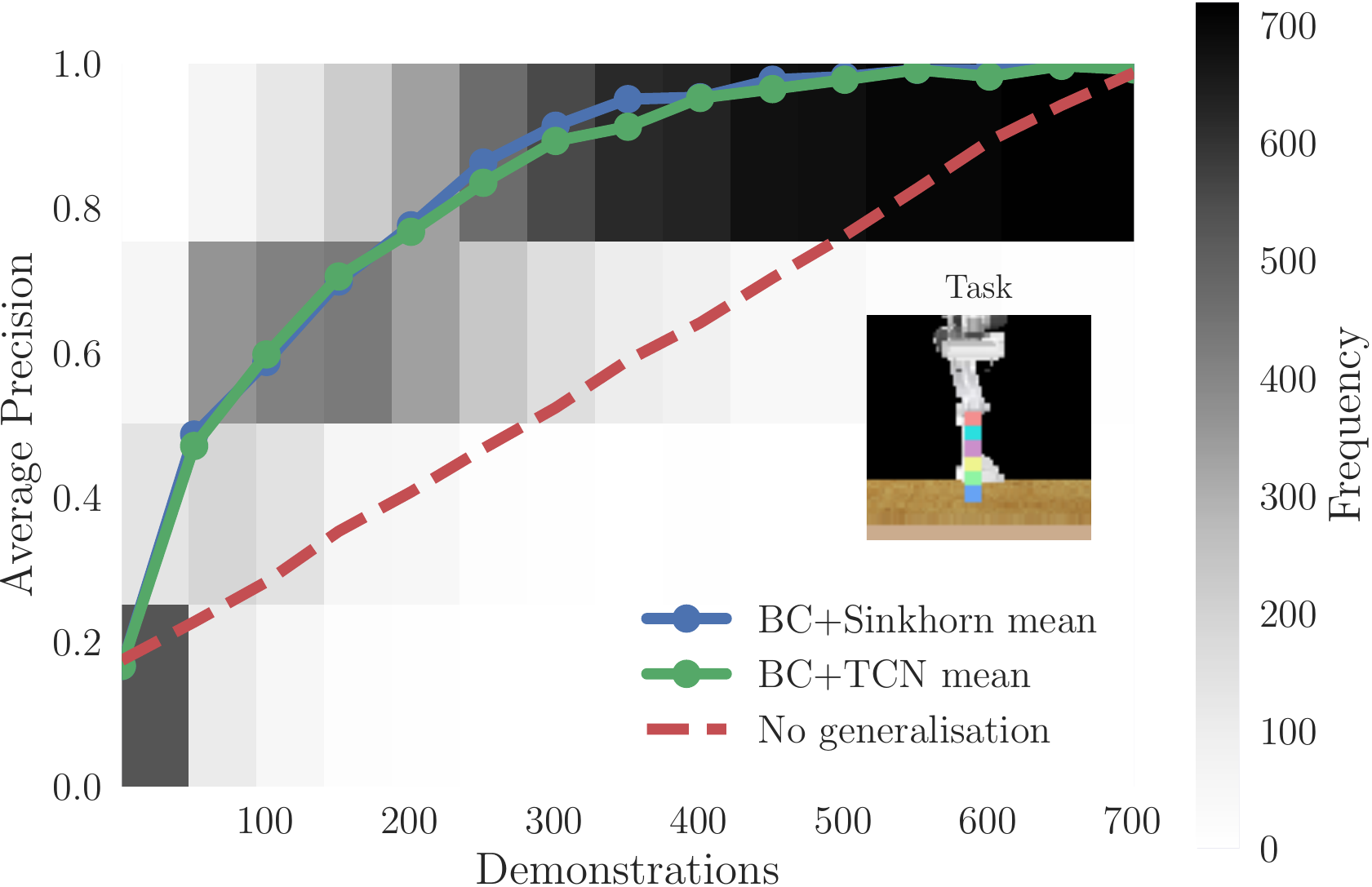}
    \caption{Both BC+Sinkhorn and BC+TCN are able to generalise to previously unseen configurations. Plots are overlaid on a histogram of average precision scores (BC+Sinkhorn) obtained for increasing numbers of demonstrations. With unique actions, models perform similarly.}
    \label{fig:generalisation}
     \vspace{-5mm}
\end{figure}
In order to investigate the reasons for performance differences between BC+TCN and BC+Sinkhorn, we explore the generalisation capabilities of the action sequencing models using a modified tower building experiment, with 6 uniquely coloured blocks (avoiding the potential for action ambiguity). We generate a single demonstration pair for each of the 720 possible tower permutations, and train models on increasing numbers of demonstrations. We then test on all 720 demonstrations. 

If the behaviour cloning models are capable of generalisation to unseen tower permutations, we would expect the precision of predicted action sequences to be better than a baseline approach of random ordering for unseen permutations and perfect ordering for previously seen permutations.

Fig. \ref{fig:generalisation} shows these results. The dashed line shows the hypothetical precision for action sequences that would be obtained by an approach that memorises previously seen action sequences and randomly guesses orders for unseen sequences. Both BC+Sinkhorn and BC+TCN networks are able to generalise to previously unseen action sequence configurations, and perform similarly in this setting.

This contrasts with the previous set of experiments and indicates that a primary advantage BC+Sinkhorn has over BC+TCN is in dealing with symmetries that arise due to action ambiguities, where more than one action can reproduce a tower, and it becomes significantly more important to reason about prior actions taken in a sequence. When there is no ambiguity in actions, which is rarely the case in robotics applications, TCNs perform similarly to Sinkhorn networks. 

\subsection{Variable length action sequencing}
\label{sec:variable_length}

We investigate variable length action sequencing using a third and final tower building experiment (with actions as in Figure \ref{fig:concept}, but with variable length action sequences -- tower heights ranging from 2 to 6 blocks). As before, all possible permutations of demonstrations are generated (1950), and models are trained on increasing numbers of demonstrations. Since variable length sequences are required, we make use of the stopping mask extension of Section \ref{sec:stopping} for the BC+Sinkhorn networks. We also compare against BC+TCN+Hungarian, extended to deal with variable length sequences through the inclusion of a stopping action class.

\begin{figure}
    \centering
    \includegraphics[width=0.46\textwidth]{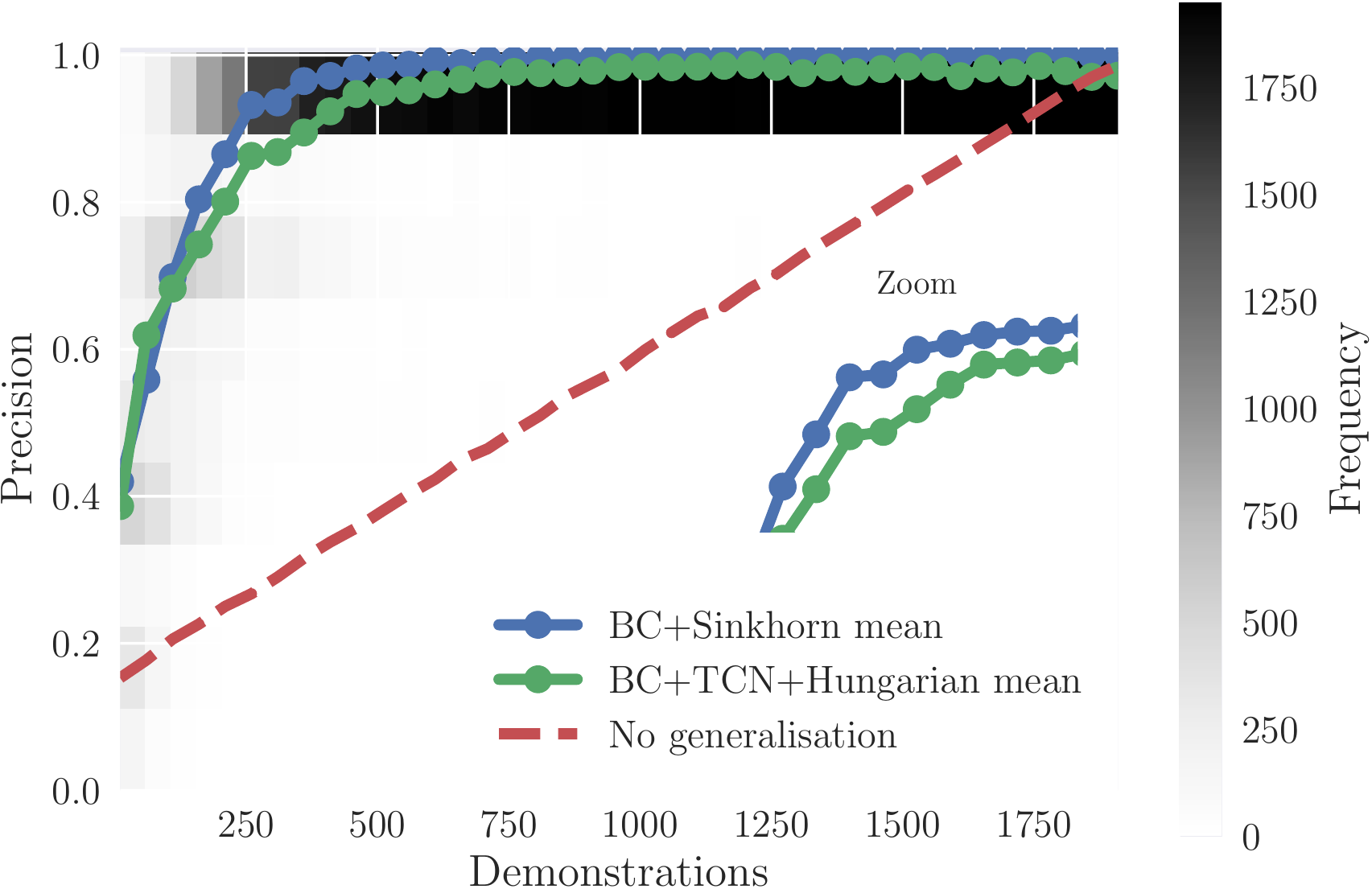}
    \caption{Action sequencing using BC+Sinkhorn shows impressive generalisation properties when action subsets are considered. Plots are overlaid on a histogram (BC+Sinkhorn) of block colour precision scores for sequences in the test set.}
    \label{fig:generalisation_subsets}
    %  \vspace{-3mm}
\end{figure}
Fig. \ref{fig:generalisation_subsets} shows these results. As before, the dashed line shows the hypothetical precision for action sequences that would be obtained by an approach that memorises previously seen action sequences and randomly guesses orders for unseen sequences. Interestingly, when training using subsets, we obtain more rapid generalisation than in the seemingly simpler case investigated earlier. This occurs because it becomes increasingly likely that action subsets have been seen within demonstrations as more training data is used. 

The task of classifying the number of actions required for a subset is relatively simple for this tower building task, and the stopping mask prediction network successfully identifies the number of actions required to build a tower after approximately 200 demonstrations have been seen. 

Importantly, the performance gains by BC+Sinkhorn over BC+TCN+Hungarian indicate that an inductive bias towards assignment assists with representation learning when there is ambiguity (colour repetitions) in the action set.

\subsection{Soma puzzle: initialising plans with sequence predictions}
\label{sec:soma}
\begin{table}
    \centering
    \caption{Soma cube results}
    \begin{tabular}{|l|c|c|}
    \hline 
         \bf{Initialisation} & \bf{Initial} & \bf{Planning iterations} \\
         & \bf{Collapses \%} & \bf{Mean $\pm$ Std. Dev.} \\
         \hline \hline
         \bf{Random} & - & $5.35 \pm 4.12$ \\
         \bf{BC+TCN+Hungarian} & $54.95 \pm 10.99$ & $1.90 \pm 3.25$\\
         \bf{BC+Sinkhorn} &  $51.55 \pm 6.19$ & $1.75 \pm 3.19$ \\
         \hline
    \end{tabular}
    \label{tab:soma}
    \vspace{-5mm}
\end{table}
\begin{figure}
    \vspace{-6mm}
    \centering
    \includegraphics[width=0.46\textwidth]{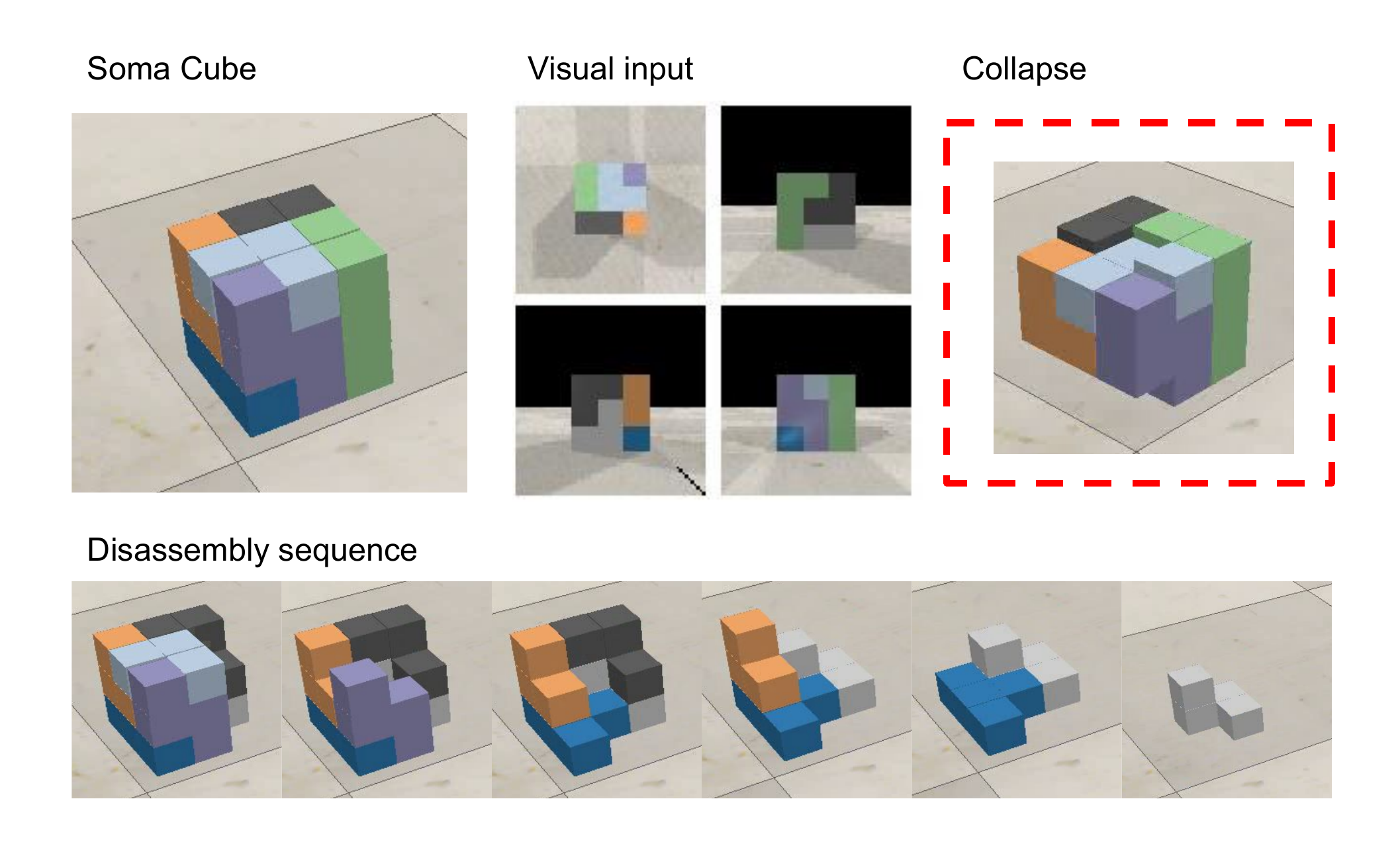}\
    \caption{Soma puzzle disassembly. Here, the task is to predict the block removal sequence in order to disassemble the Soma cube, given a set of four input images of the cube, captured from different sides. Failure to correctly predict the removal  sequence will result in the cube collapsing, placing the environment in a state where pre-scripted action sequences can no longer be used.}
    \label{fig:soma}
    \vspace{-5mm}
\end{figure}

\begin{figure*}
    \centering
    \begin{minipage}{.225\textwidth}
    \includegraphics[width=\textwidth]{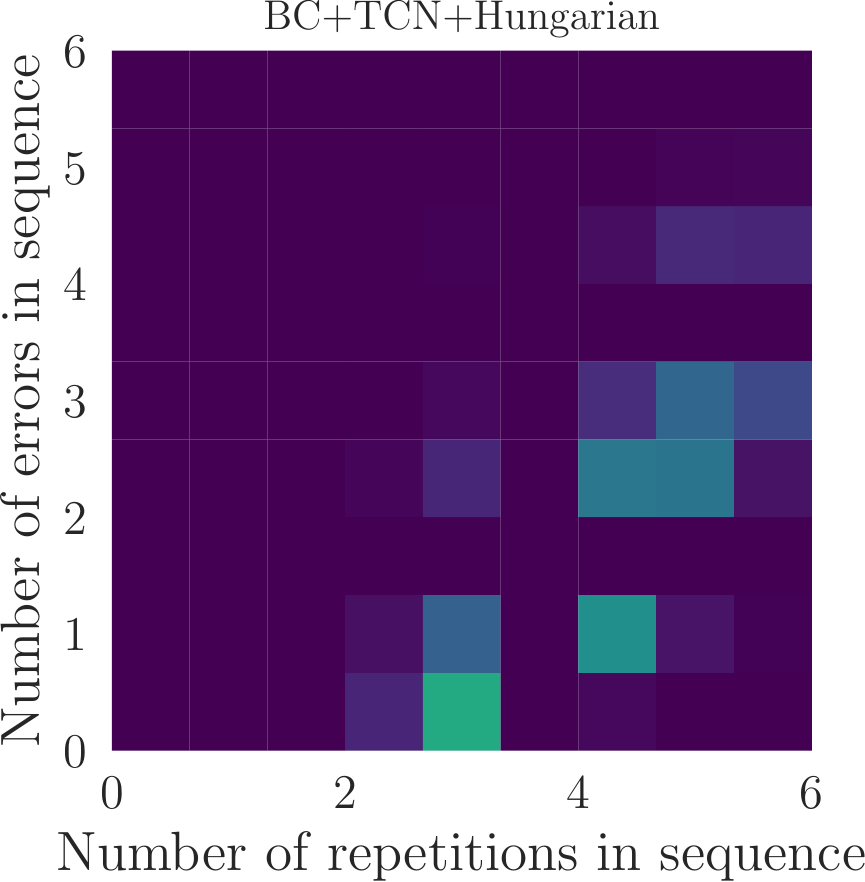}
    \end{minipage}
    \begin{minipage}{.225\textwidth}
    \includegraphics[width=\textwidth]{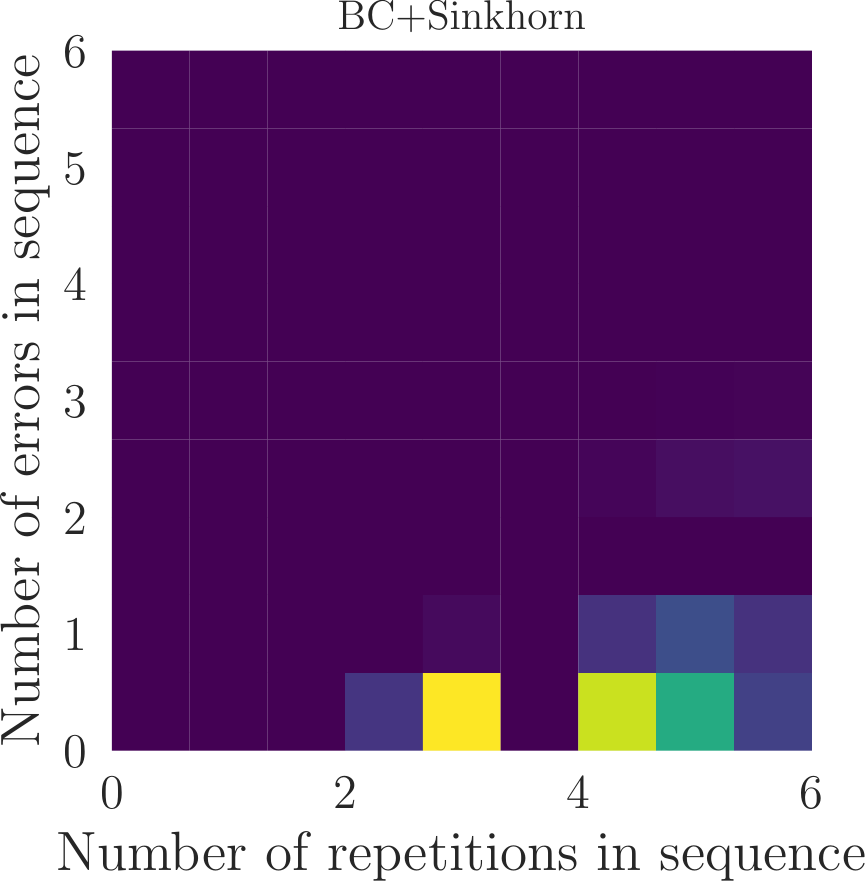}
    \end{minipage}
    \begin{minipage}{0.05\textwidth}
    \includegraphics[width=\textwidth]{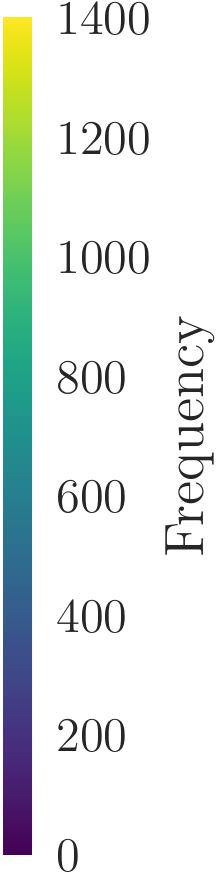}
    \end{minipage}
    \begin{minipage}{0.48\textwidth}
    \includegraphics[width=\textwidth]{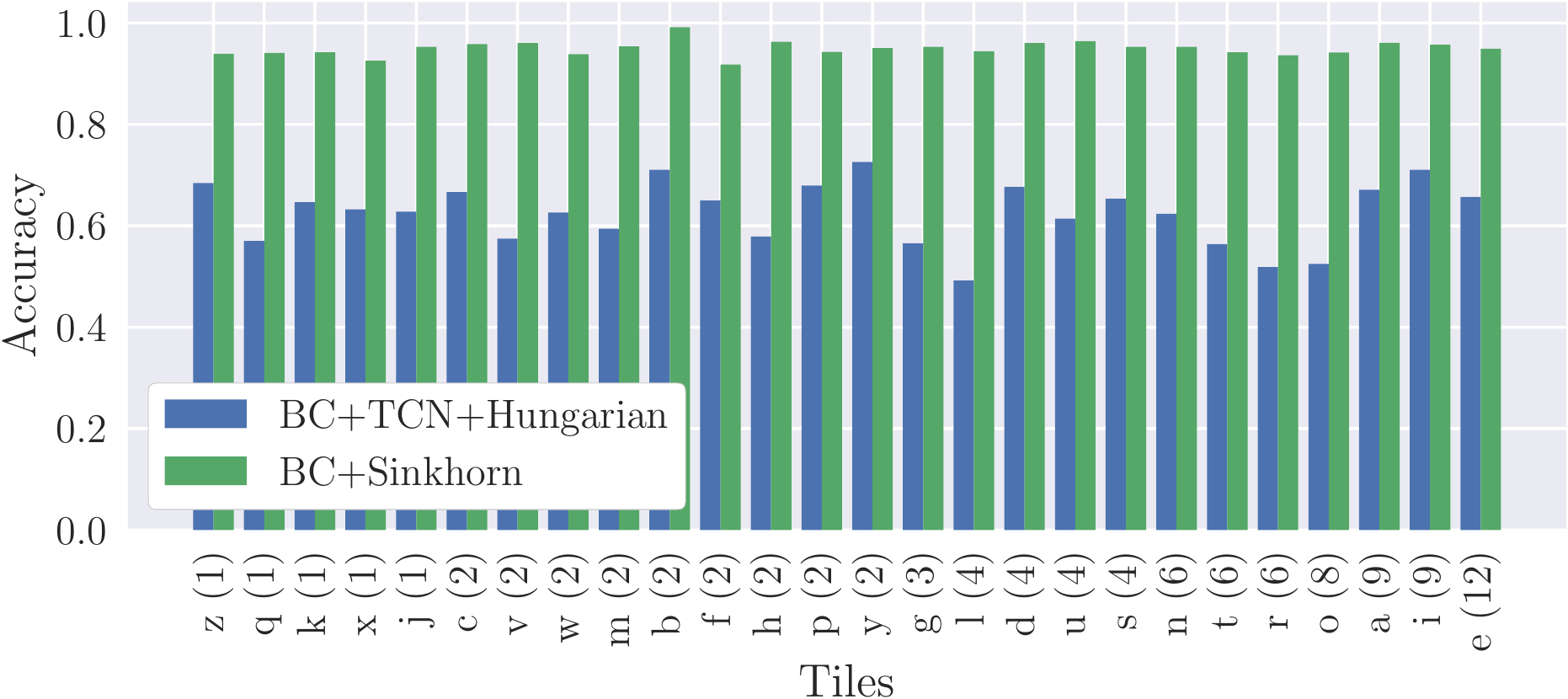}
    \end{minipage}
    \caption{Tile recognition error increases for TCN's as the number of tile repetitions in the test sequence increases, but the failures themselves are spread relatively evenly across tiles when tested using the full scrabble tile set. Bracketed numbers indicate frequency of occurrence in the tile set.}
    \label{fig:tile_error}
    \vspace{-2mm}
\end{figure*}

\begin{figure}
    \centering
    \includegraphics[width=0.45\textwidth]{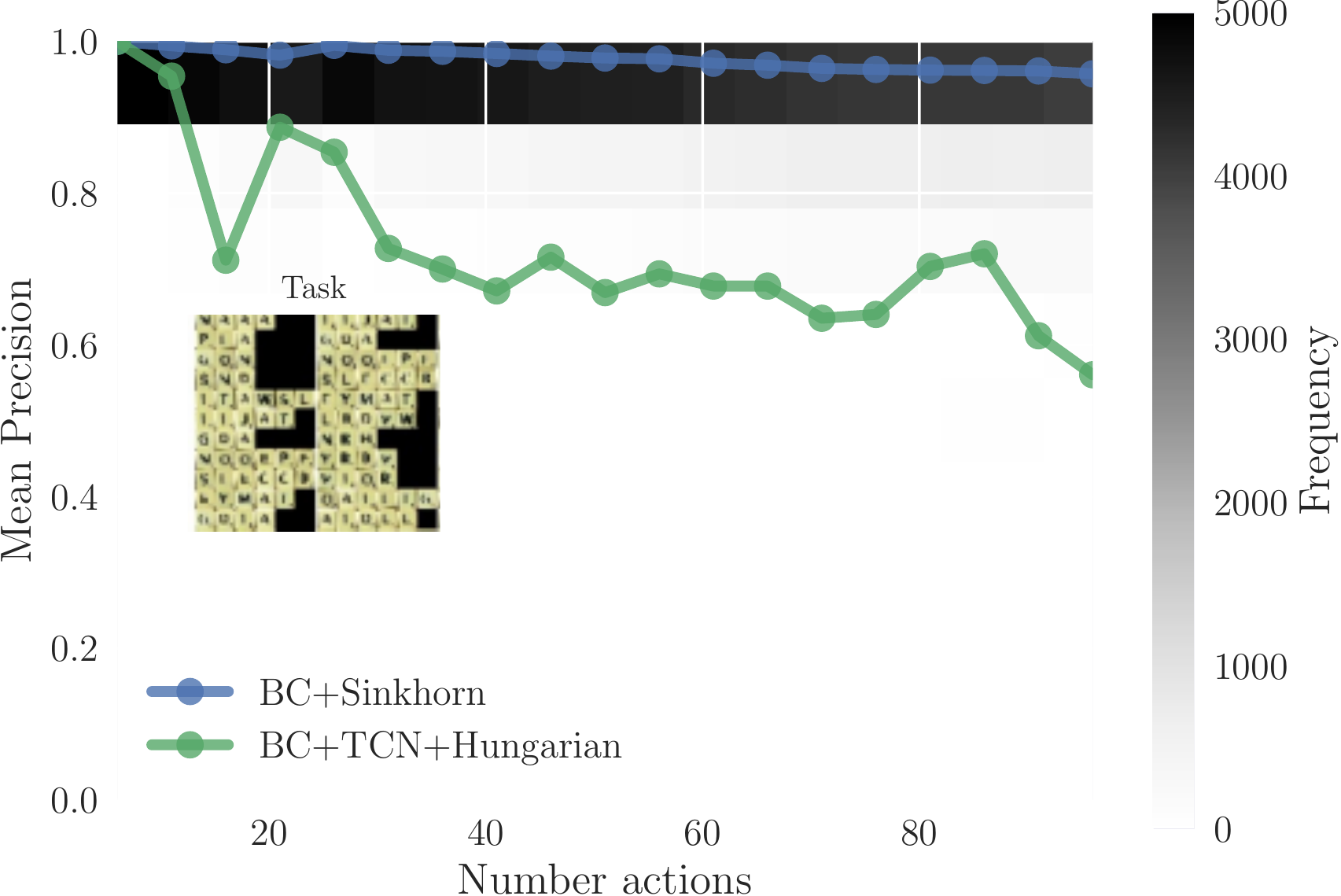}
    \caption{Performance as action set size increases.}
    \label{fig:generalisation_scale}
    \vspace{-5mm}
\end{figure}

The ability to reason about action sequence permutations is  particularly important in assembly or disassembly tasks. We investigate the ability of behaviour cloning to solve more complex tasks, using a Soma puzzle \cite{soma}. As illustrated in Fig. \ref{fig:soma}, Soma puzzles consist of 7 distinctly shaped blocks, which can be assembled into arbitrary shaped objects. Here, we consider the task of disassembling a 3x3 Soma cube, which can be constructed in 240 distinct ways (ignoring reflections and rotations). 

The Soma puzzle has a long history in robotics and robot learning \cite{MALCOLM1990123,malcolm1995somass}, and has been the study of extensive research due to the complexity of shapes that can be constructed with it. The geometry of the puzzle parts means that disassembling the puzzle using pre-scripted actions requires that parts be removed in a precise order. Failure to do so will result in the puzzle collapsing, placing the environment in a state where pre-scripted actions can no longer be used. Correctly predicting part extraction order from images of the cube is challenging as it requires a model that can reason about how parts interlock and their relative positioning. 

This problem is also a challenge for traditional planning algorithms, requiring a careful, and non-trivial, specification of relationships between components and problem constraints and a backtracking search over numerous possible action sequences. To investigate this setting, models were trained using a dataset comprising images of the 240 possible initial puzzle configurations, and a manually defined extraction order for each puzzle, randomly (repeated 100 times with different seeds) split into 120 training and validation examples, and 120 test examples. 

As shown in Table \ref{tab:soma}, despite outperforming the temporal convolutional architecture (BC+TCN+Hungarian), Sinkhorn behaviour cloning (BC+Sinkhorn) is still only successful on about half of the test cases when predicted action sequences are directly applied. However, when the predicted action sequences are used to initialise a suitable planning algorithm\footnote{A backtracking search using the simulator in the loop to test for failures.} there are substantial gains in planning time, with a clear reduction in the number of iterations used to search for a suitable planning order. This ability to speed up symbolic planning using surrogate neural models is particularly valuable in more general robot assembly tasks. 

In this case, there is a small performance difference between the TCN and Sinkhorn models. However, as will be shown next, results on larger actions sets indicate that Sinkhorn networks scale far better than TCNs, and these performance differences become substantially more pronounced as more actions are considered.

\subsection{Scrabble: scaling to larger action sets}
\label{sec:scale}

% \begin{figure}
%     \centering
%     \includegraphics[height=0.195\textwidth]{./figs/Error_analysis_tcn-crop}   \includegraphics[height=0.195\textwidth]{./figs/Error_analysis_sink-crop}
%     \caption{Test results for the full Scrabble tile set show that TCNs make more errors for sequences with more actions repetitions.}
%     \label{fig:errors_ambiguity}
% \end{figure}

A simplified Scrabble setting is used to evaluate the ability of BC+Sinkhorn to scale to larger action sets. Here, a standard English Scrabble tile set is used to generate images (10000) of random letter combinations with lengths 3 to 6, sampled from an increasingly large subset of the full tile set. We test on a set of randomly generated test words (5000).

The scrabble setting is also useful as it shows how permutations could be used when an action needs to be used more than once, as certain letters can be used multiple times.

An analysis of the prediction errors made by the TCN (see Fig. \ref{fig:tile_error}) shows consistent performance degradation across all letters, regardless of frequency of repetition, but that failures are more likely in sequences with more character repetitions.

Fig. \ref{fig:generalisation_scale} shows the decrease in performance (mean average spelling precision) as the number of actions used to generate training and test data is increased. BC+Sinkhorn shows extremely impressive scalability, with only small decreases in performance as the action set size increases. This could be remedied by additional training data, although training becomes time consuming with larger action sets\footnote{As the number of actions increased, we observed that we needed to train for substantially longer before reaching convergence, with our largest model (98 actions) requiring approximately 5000 epochs to converge.}. In contrast BC+TCN+Hungarian networks become increasingly unreliable as the action set size is increased, with significant performance drops.

By virtue of the inductive bias for permutations, Sinkhorn networks provide a more principled way of dealing with action ambiguities in the architecture, forcing the network to rule out/select actions by taking into account other actions that are taken in the sequence. This appears to produce representations that are better equipped to deal with this ambiguity, and results in improved performance.

\section{CONCLUSION}
\label{sec:conclusion}

This paper introduces a permutation prediction approach to vision-based neural action sequencing. Action sequencing using latent permutations predicted by Sinkhorn networks is most effective in tasks where there are potentially multiple actions leading to a desired state, and where there are constraints on the number of times an action can be used. Results show that neural action sequencing provides valuable improvements in planning speed when used to initialise planning algorithms, and experiments showed that impressive generalisation can be obtained using these networks. Importantly, Sinkhorn networks are able to scale to far greater action set sizes than temporal convolution networks. Temporal convolution and Sinkhorn networks are similar capacity models, and there are no major computational differences in the forward pass, which means that latent permutations are a promising and useful approach to surrogate sequence modelling for planning enhancement in robotics.

\section*{APPENDIX}

% \section*{Experimental settings}

See below for a summary of experimental settings. The accompanying \href{https://youtu.be/F86hHFQDGW4}{video} illustrates failure modes and example use cases.

\subsection*{Tower building}
\vspace{-2mm}
\begin{figure}[!ht]
    \centering
    \includegraphics[width=0.48\textwidth]{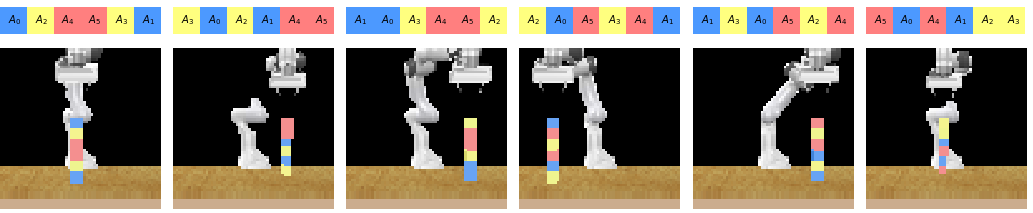}
    \caption{Training data for tower building experiments consists of pick and place action sequences and the corresponding images of completed towers. Action sequences are coloured in accordance with the block they correspond to moving for visual clarity. Our goal is to predict the correct sequence of actions required to reproduce an input image scene.}
    \label{fig:demo_data}
\end{figure}

\vspace{-2mm}
\begin{wraptable}{l}{0.22\textwidth}
\centering
\vspace{-4mm}
\begin{tabular}{|l|}
\hline
    CNN Encoder\\
\hline
     32 5x5 kernels, ReLU\\
     64 5x5 kernels, ReLU\\
     2x2 MaxPool2D\\
     128 5x5 kernels, ReLU\\
     2x2 MaxPool2D\\
     256 5x5 kernels, ReLU\\
     2x2 MaxPool2D\\
     Dropout (p=0.5)\\
     128 Neuron FC, ReLU\\
     \hline
     \end{tabular}
     \vspace{-4mm}
\end{wraptable}

Six blocks (2 blue, 2 yellow, 2 red) were used for tower building (baseline and subsets experiments) in CoppeliaSim, with primitive actions to pick up a block from a pre-defined start position and place it above the last placed block (no action is taken to place the first block in the sequence). 6 uniquely coloured blocks were used for ablation generalisation experiments.

Both Sinkhorn and TCNs used a CNN encoder with parameters listed to the left. Behaviour cloning and Sinkhorn networks were trained with batch sizes of 16 for 10000 epochs, while the TCNs were trained for 2000 (all models were trained until convergence, but we found TCNs converged far faster than alternative models), both using Adam \cite{kingma2014adam} and a learning rate of 3e-4. TCNs used 6 layers of length 6, the length of the maximum action set size.

\begin{figure}
\centering
\includegraphics[height=0.2\textwidth]{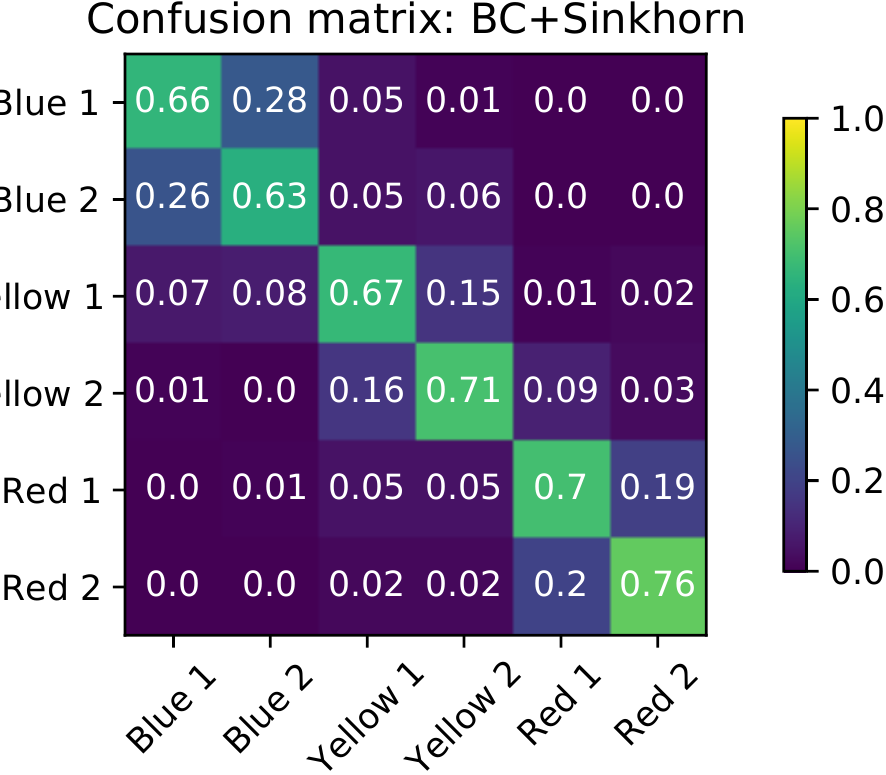}
\includegraphics[height=0.2\textwidth]{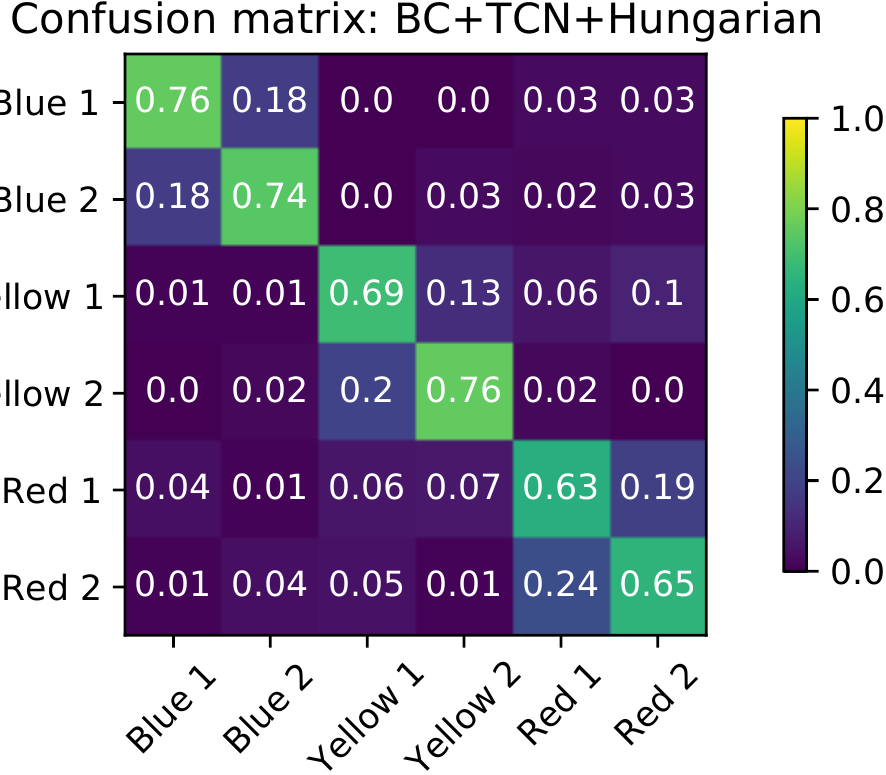}
\caption{Confusion matrices for a tower model trained using 200 examples indicate that models with the latent permutation inductive bias better associate visual input with actions than TCNs.}
\vspace{-6mm}
\end{figure}
    
\subsection*{Soma puzzle}

% \begin{figure}[!ht]
%     \centering
%     \includegraphics[width=0.4\textwidth]{./figs/soma_pieces.png}
%     \caption{Soma puzzle pieces.}
%     \label{fig:my_label}
% \end{figure}
Soma puzzles consist of 7 distinctly shaped blocks, which can be assembled into arbitrary shaped objects. Here, we consider the task of disassembling a 3x3 Soma cube, which can be constructed in 240 distinct ways (ignoring reflections and rotations). Soma solutions were obtained using Polyform Puzzler, \url{http://puzzler.sourceforge.net}, a set of solvers for polyforms.

For Soma puzzle extraction planning, a backtracking search was used to find collapse free extraction orders. Here, actions that trigger collapse were randomly swapped with later actions, using the simulator in the loop to identify collapses. Experiments compared planning speed when search was initialised using a random starting sequence and using sequence predictions from the neural networks.

Soma puzzle parts were modelled in PyRep \cite{james2019pyrep} using independent cubes, and collapse detection was implemented by checking if any of the cubes had moved after a part had been deleted from the scene.

The same CNN encoder architecture used for tower building was used here, but models were trained with batch sizes of 32, using Adam and a learning rate of 3e-4.

\subsection*{Scrabble}

\vspace{-2mm}
\begin{wrapfigure}{l}{0.22\textwidth}
\vspace{-3mm}
\centering
\includegraphics[height=0.13\textwidth]{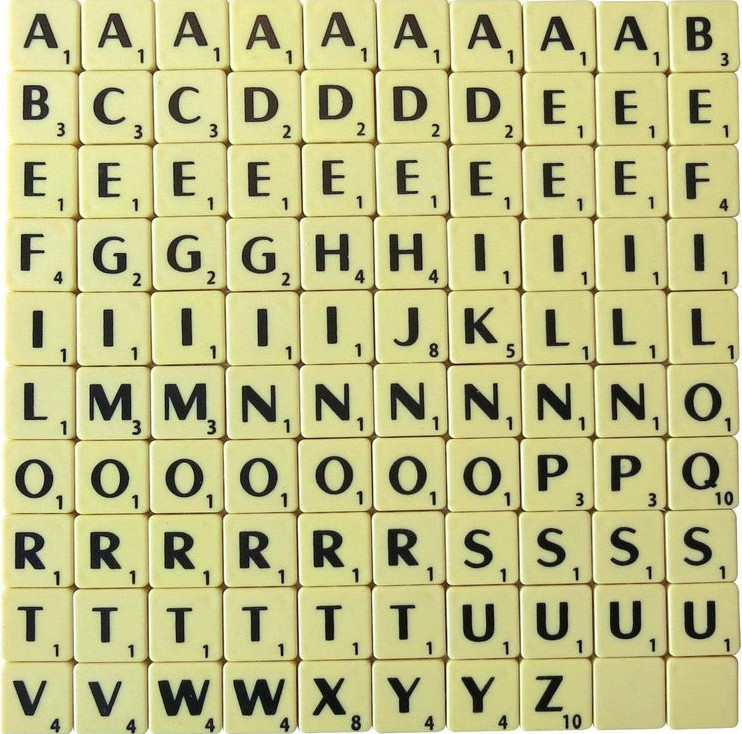}\includegraphics[height=0.13\textwidth]{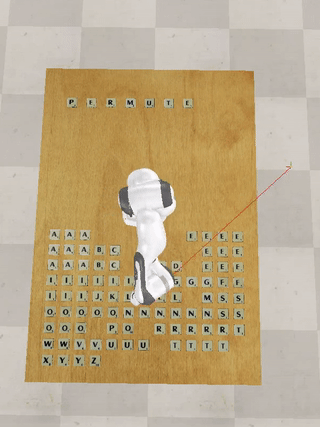}
\caption{Tile set used for image generation and demonstration in Coppeliasim.}
\vspace{-4mm}
\end{wrapfigure}A standard English scrabble tile set (98 possible tiles without blanks) is used to generate images of randomised letter combinations. Letters can be used multiple times, which introduces additional complexity when grounding image components and actions. Images are restricted to 6 characters or fewer, due to resolution constraints.
%-- 98 possible tiles comprising alphabet letters with frequencies a=9, b=2, c=2, d=4, e=12, f=2, g=3, h=2, i=9, j=1, k=1, l=4, m=2, n=6, o=8, p=2, q=1, r=6, s=4, t=6, u=4, v=2, w=2, x=1, y=2, z=1 -- 

Both BC+TCN and BC+Sinkhorn used a Resnet18 encoder \cite{he2016deep} for input images and were trained using Adam with a learning rate of 1e-4. Models were trained with a batch size of 64. TCNs used 6 temporal convolution layers of length 7, the maximum number of actions in an action sequence. Sinkhorn networks used a fixed size latent bottleneck state of 128 dimensions, while TCNs used 7 latent states of 16 dimensions each. Both models were trained for 5000 epochs.

% Robot demonstrations were conducted in CoppeliaSim using PyRep, using pre-scriped actions, pick tile $k$, place at position $x + \delta_i$, where $\delta_i$ denotes an offset corresponding to the $i$-th action in a sequence.

% \section*{ACKNOWLEDGMENT}

%  We are grateful to Yordan Hristov and members of the Robust Autonomy and Decisions Group for discussions and feedback.

\bibliographystyle{IEEEtran} 
\bibliography{root}

% \addtolength{\textheight}{-18cm} 

\end{document}